%% file: colm2024_conference.tex
\documentclass[table]{article}
\PassOptionsToPackage{table}{xcolor}

\usepackage{colm2024_conference}

\usepackage{booktabs}
\usepackage{graphicx}
\usepackage{enumitem}
\usepackage{wrapfig}
\usepackage{algorithm}
\usepackage{algpseudocode}
\usepackage{multicol}

\usepackage{microtype}
\usepackage{amsmath}
\usepackage{colortbl}
\usepackage[utf8]{inputenc}
\definecolor{lightgray}{rgb}{0.9,0.9,0.9}
\usepackage{caption}
\usepackage{subcaption}
\usepackage{xcolor}
\usepackage{setspace}
\usepackage{url}
\usepackage{multirow}
\usepackage{colortbl}
\usepackage{tabularx}
\usepackage{blindtext}
\usepackage{pgfplots}
\pgfplotsset{compat=1.18} 
\usepackage{tikz}
\usetikzlibrary{er,positioning,bayesnet}
\usepackage{makecell}
\usepackage{tipa}
\usepackage{siunitx}
\usepackage{nicefrac}
\usepackage{tocloft}
\usepackage{listings}
\usepackage[raster,skins]{tcolorbox} %
\usepackage{xltabular}
\usepackage{adjustbox}
\usepackage{xurl}
\usepackage{setspace}

\usepackage{lipsum}

\usepackage{arydshln} 
\usepackage[table]{xcolor}
\usepackage{varwidth}

\input{math_commands.tex}

\title{Hunyuan-TurboS: Advancing Large Language Models through Mamba-Transformer Synergy and Adaptive Chain-of-Thought}

\author{
\bf Tencent Hunyuan Team
}

\begin{document}

\maketitle

\begin{abstract}

As Large Language Models (LLMs) rapidly advance, we introduce Hunyuan-TurboS, a novel large hybrid Transformer-Mamba Mixture of Experts (MoE) model. It synergistically combines Mamba's long-sequence processing efficiency with Transformer's superior contextual understanding. Hunyuan-TurboS features an adaptive long-short chain-of-thought (CoT) mechanism, dynamically switching between rapid responses for simple queries and deep "thinking" modes for complex problems, optimizing computational resources. Architecturally, this 56B activated (560B total) parameter model employs 128 layers (Mamba2, Attention, FFN) with an innovative AMF/MF block pattern. Faster Mamba2 ensures linear complexity, Grouped-Query Attention minimizes KV cache, and FFNs use an MoE structure. Pre-trained on 16T high-quality tokens, it supports a 256K context length and is the first industry-deployed large-scale Mamba model. Our comprehensive post-training strategy enhances capabilities via Supervised Fine-Tuning (3M instructions), a novel Adaptive Long-short CoT Fusion method, Multi-round Deliberation Learning for iterative improvement, and a two-stage Large-scale Reinforcement Learning process targeting STEM and general instruction-following. Evaluations show strong performance: overall top 7 rank on LMSYS Chatbot Arena with a score of 1356, outperforming leading models like Gemini-2.0-Flash-001 (1352) and o4-mini-2025-04-16 (1345). TurboS also achieves an average of 77.9\% across 23 automated benchmarks. Hunyuan-TurboS balances high performance and efficiency, offering substantial capabilities at lower inference costs than many reasoning models, establishing a new paradigm for efficient large-scale pre-trained models.

\end{abstract}

\vfill

\begin{figure}[hbp]
    \centering
    \includegraphics[width=0.99\textwidth]{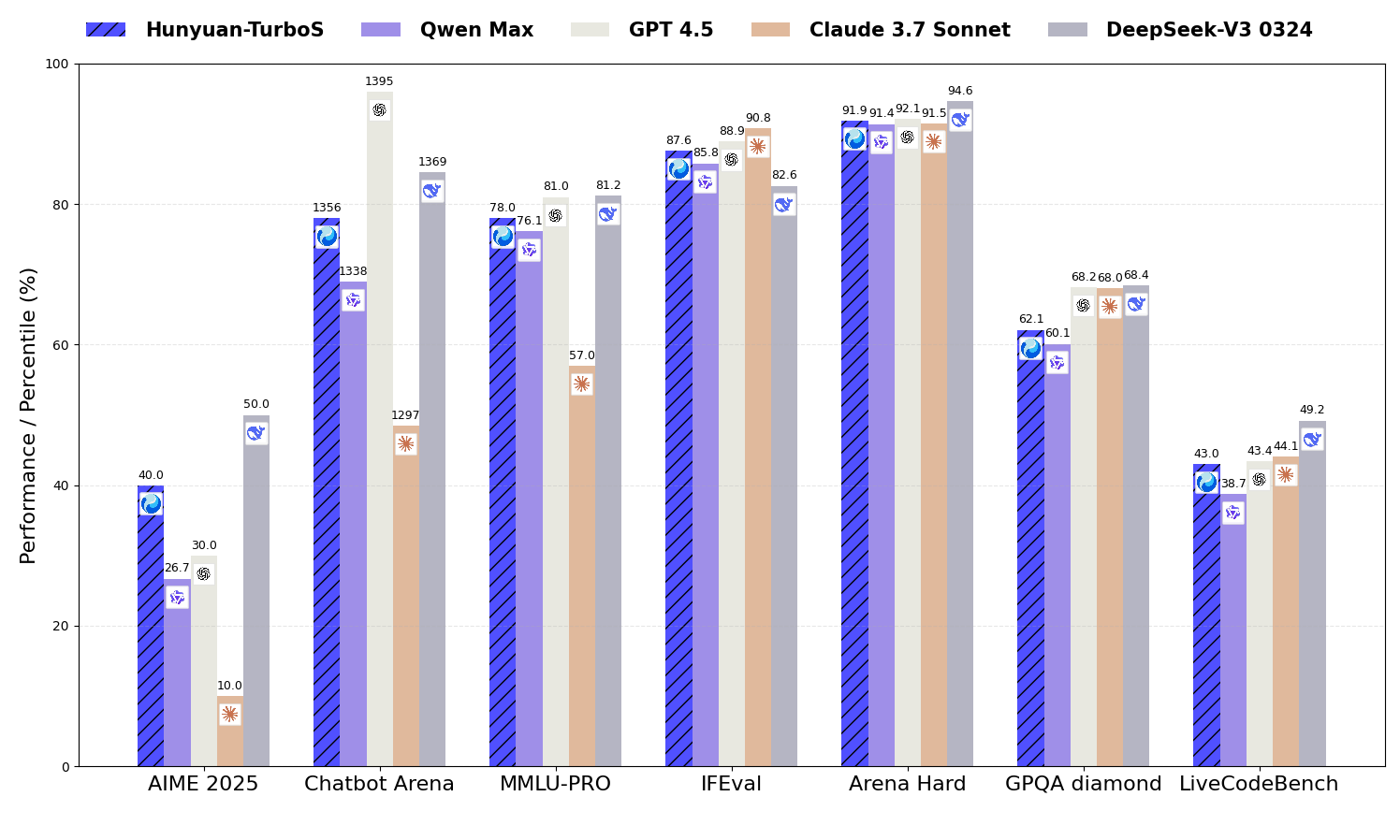}
    \caption{Benchmark performance of Hunyuan-TurboS.}
    \label{fig:intro}
\end{figure}

\vfill

\input{content/intro.tex}

\input{content/pretraining}

\input{content/posttrain}

\input{content/experiments.tex}

\input{content/arch.tex}

\input{content/conclusion.tex}
\input{content/authors.tex}

\clearpage
\bibliography{biblio}
\bibliographystyle{colm2024_conference}

\clearpage
\appendix
\input{content/appendix.tex}

\end{document}

%% file: math_commands.tex
\usepackage{amsmath,amsfonts,bm}

\def\eqref#1{equation~\ref{#1}}

\def\1{\bm{1}}

\DeclareMathAlphabet{\mathsfit}{\encodingdefault}{\sfdefault}{m}{sl}
\SetMathAlphabet{\mathsfit}{bold}{\encodingdefault}{\sfdefault}{bx}{n}



%% file: content/intro.tex
\section{Introduction}
\label{sec:intro}

Large Language Models (LLMs) have witnessed unprecedented acceleration in their development, rapidly advancing towards artificial general intelligence (AGI) through recent breakthroughs in foundation models. Advanced systems such as GPT-4o~\citep{gpt4o}, Gemini 2.5~\citep{gemini2.5}, DeepSeek-R1~\citep{deepseek_r1}, and Qwen3~\citep{yang2025qwen3} now demonstrate capabilities that significantly narrow the gap between specialized AI and the generalized intelligence humans have long sought to create. Aiming to further push these boundaries, we introduce Hunyuan-TurboS, a large hybrid Transformer-Mamba Mixture of Experts (MoE) model.

Hunyuan-TurboS exhibits several innovative characteristics that establish it as a powerful LLM achieving an excellent balance between performance and efficiency. First, it synergistically combines the efficient long-sequence processing capabilities of the Mamba~\citep{gu2023mamba} architecture with the superior contextual understanding of the Transformer~\citep{vaswani2017attention} architecture. Second, it employs an adaptive long-short chain-of-thought mechanism. This integrates the advantages of short chain-of-thought models (e.g., GPT-4o~\citep{gpt4o}), such as rapid response and computation-friendly inference, with the excellent complex reasoning capabilities of long chain-of-thought models (e.g., o3~\citep{o3}). When faced with simple questions, TurboS automatically activates a "no thinking" mode to deliver results of sufficient quality at minimal computational cost. Conversely, when encountering complex problems, TurboS automatically switches to a "thinking" mode, employing deep reasoning methods such as step-by-step analysis, self-reflection, and backtracking to arrive at highly accurate answers.

Hunyuan-TurboS is a hybrid architecture that integrates Transformer, Mamba2 \citep{dao2024transformers}, and Feed-Forward Network (FFN) components, designed for scalability and efficiency in both training and inference. With 128 layers (57 Mamba, 7 Attention, and 64 FFN), the model scales to 56B activated parameters and 560B total parameters. The architecture employs an "AMF" (Attention → Mamba2 → FFN) and "MF" (Mamba2 → FFN) block pattern to balance performance and efficiency, for long-context tasks. Mamba2 layers achieve linear complexity, while Grouped-Query Attention (GQA) \citep{ainslie2023gqatraininggeneralizedmultiquery} further minimizes KV cache overhead. The FFN layers use an MoE structure with 32 experts, activating 1 shared and 2 specialized experts per token. Hunyuan-TurboS is pre-trained on a 16T high-quality token dataset, supporting a context length of up to 256K. Notably, Hunyuan-TurboS is the first industry-deployed large-scale Mamba-based model, setting a new paradigm for efficient large-scale pre-trained models.

Our post-training strategy for Hunyuan-TurboS encompasses four critical modules designed to significantly enhance its capabilities. Initially, Supervised Fine-Tuning establishes a robust foundation by methodically curating 3M natural and synthetic \citep{wang2022self, luo2023wizardcoder, zeng2024automatic,luo2023wizardmath,wei2023magicoder} instruction data, categorized by comprehensively delineated topics across diverse domains and implementing multi-dimensional metrics for rigorous filtering and quality assurance. We then introduce a novel Adaptive Long-short Chain-of-Thought Fusion method, enabling the model to autonomously select optimal reasoning strategies, efficiently allocate computational resources, and enhance response readability through lossless compression and reformatting of lengthy chains of thoughts, achieved via a teacher model refined through dedicated SFT and a unique reinforcement learning framework with difficulty-adaptive and CoT compression rewards. Subsequently, Multi-round Deliberation Learning involves the SFT model comparing other cutting-edge Hunyuan models in a simulated evaluation environment, an iterative refinement cycle driven by evaluations from a multi-LLM judge ensemble and human expert oversight to strategically identify and address capability gaps, informing subsequent SFT iterations. Finally, a Two-stage Large-scale Reinforcement Learning process, leveraging GRPO, further hones the model. The first stage focuses on bolstering reasoning capabilities, while the second stage aims to improve general instruction-following proficiency across all domains. This synergistic, multi-faceted strategy aims to cultivate a highly capable, efficient, and adaptable language model.

Both human and automated evaluations demonstrate the competitive capabilities of Hunyuan-TurboS. In the LMSys Chatbot Arena, known for its blind, side-by-side human evaluations that minimize bias and offer objective capability assessment, Hunyuan-TurboS achieved a notable score of 1356. This places it among the top 7 models, outperforming strong contenders like Gemini-2.0-Flash-001, o4-mini and Gemma-3-27B-it. Notably, in categories such as Math, Multi-Turn, and Longer Query, our model ranks among the top 5. For comprehensive automated assessment, we evaluated Hunyuan-TurboS across 23 benchmarks encompassing mathematical reasoning, logical reasoning, code generation, knowledge, alignment tasks, and instruction following. The model achieved an average score of 77.9\%, significantly outperforming comparable non-reasoning models. While its performance approaches that of computationally intensive reasoning models, Hunyuan-TurboS demonstrates a remarkable balance of efficiency and effectiveness, particularly considering its substantially lower inference costs.

\newpage

Our main contribution includes:
\paragraph{Pre-Training:}

\begin{enumerate}[label=(\arabic*)]
    \item \textbf{Data Recipe:} We developed a meticulous data curation pipeline for 16T high-quality tokens, improving quantity, quality, and diversity through systematic filtering, deduplication, and specialized content extraction modules.
    \item \textbf{560B Mamba:} We designed a novel 560B total parameter hybrid Transformer-Mamba2-MoE architecture, featuring innovative AMF/MF block patterns. This synergizes Mamba2's efficiency with Transformer's contextual understanding.
    \item \textbf{Annealing:} We implemented multi-stage pre-training refinements including an annealing phase with diverse data and curriculum-based long-context extension up to 256K tokens using NTK-aware positional encoding.
\end{enumerate}

\paragraph{Post-Training:}

\begin{enumerate}[label=(\arabic*)]
    \item \textbf{Data Recipe:} We curated a 3M instruction dataset for robust Supervised Fine-Tuning, categorized by diverse topics with rigorous quality assurance, establishing strong foundational capabilities for the model.
    \item \textbf{Adaptive CoT Fusion:} We developed an Adaptive Long-short Chain-of-Thought Fusion method, enabling dynamic selection of reasoning strategies via a specially trained teacher model and reinforcement learning. Notably, our model delivers performance comparable to that of top-tier reasoning model in the LMSYS Chatbot Arena while utilizing only about 50\% of the generation tokens, highlighting significant improvements in token efficiency which directly translate to reduced generation costs. 
    \item \textbf{Deliberation Learning:} We implemented Multi-round Deliberation Learning, where the SFT model iteratively refines its capabilities by competing against other models, guided by an LLM-judge ensemble and Human expert oversight.
    \item \textbf{Two-stage GRPO:} We employed a two-stage Large-scale Reinforcement Learning (GRPO) process, first targeting STEM reasoning and then general instruction-following, guided by a comprehensive general reward system.
\end{enumerate}

\paragraph{Infrastructures:}
\begin{enumerate}[label=(\arabic*)]
    \item \textbf{Angel-RL:} We built Angel-RL, an efficient reinforcement learning framework integrating training and inference, incorporating comprehensive parallelism (TP, PP, EP, CP) and innovative state-passing for context parallelism.
    \item \textbf{Mamba MoE:} We optimized inference via AngelHCF, featuring Mamba kernel enhancements (prefill/decode), MoE expert parallelism, and fp32 precision for Mamba states to improve long-text generation quality. It ultimately achieving a 1.8x speedup compared to Hunyuan-Turbo, which is a pure Transformers MoE model.
\end{enumerate}

\paragraph{Summary of Core Evaluation Results:}

\begin{enumerate}[label=(\arabic*)]
    \item \textbf{LMSYS Chatbot Arena:} Hunyuan-TurboS achieved a 1356 Arena score, ranking top 7 overall. It excelled with top 1 in Chinese, French, and Spanish, and top 5 in tasks like Hard Prompts, Creative Writing, Multi-Turn, and Longer Queries.
    \item  \textbf{Automatic Evaluations:} The model demonstrated strong performance across 23 automated benchmarks, achieving an average of 77.9\%, with notable results in mathematics, coding and general domains.
    \item \textbf{Balance efficiency and performance:} Our model effectively balances high performance and computational efficiency, delivering substantial capabilities comparable to larger reasoning models but at significantly lower inference costs across diverse evaluations.
\end{enumerate}

%% file: content/pretraining.tex
\newpage
\section{Pre-Training}
In this section, we will introduce the details of our pre-training stage in Hunyuan-TurboS, including
(a) data for pre-training, which provides a systematic pipeline for data quality control and data mix, providing fundamental information for the capability acquisition of LLMs,
(b) model structure, which proposes a novel hybrid Transformer-Mamba structure for effective and efficient LLM training and serving, with comprehensive design motivations and details,
and (c) annealing and long-context pre-training recipes, with several insights during these two essential stages in pre-training.
These techniques build the foundation of our Hunyuan-TurboS's remarkable capability to facilitate downstream applications.

\subsection{Data for Pre-training}

Pre-training data are the fundamental fuel of LLMs. Compared to Hunyuan-Large~\citep{sun2024hunyuan}, we enhance the data used for the pre-training stage in three key dimensions: quantity, quality, and diversity.

\begin{figure}[!htp]
\centering
\includegraphics[width=1.0\textwidth]{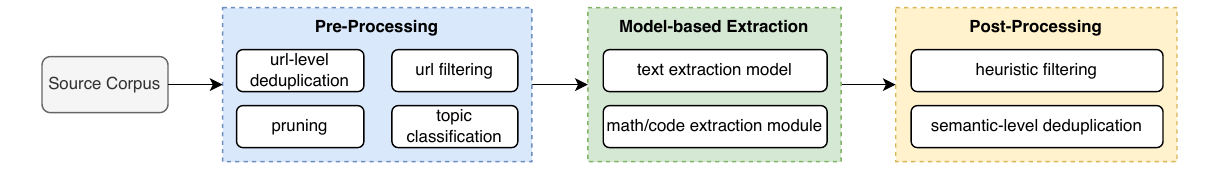}
\caption{The curation pipeline of pre-training data in Hunyuan-TurboS.}
\label{fig:data}
\end{figure}

We rigorously refine our curation pipeline to efficiently process diverse raw data sources, as illustrated in Figure~\ref{fig:data}.
The process begins with URL-level deduplication and filtering of the varied-format source data to eliminate redundant items.
Additionally, we utilize precise block-level pruning techniques to remove noisy content within each instance, further reducing the volume of low-quality data and accelerating downstream processes.
Each piece of raw content is then annotated with a specific topic label by our topic classification model, which facilitates subsequent processing stages.
Following this, a content extraction model processes the previously cleaned data to extract plain text.
To ensure the acquisition of high-quality STEM and code data, we construct several domain-specific extraction modules. 
Additionally, we employ comprehensive heuristic filtering methods to discard low-qualified extracted content, and implement global-scale semantic-level deduplication to produce the final data.

Furthermore, we optimize our quality and diversity assurance systems by developing comprehensive critique models and data mixture models. 
We introduce several foundational quality criteria with dozens of well-organized domain type labels into our critique models, enabling principled data selection and integration.
Across different phases of pre-training, we employ effective mixture models to provide varied data mixture recipes that ensure optimal data utilization.

Overall, Hunyuan-TurboS is trained on a corpus comprising $16$ trillion tokens in our tokenizer, which is the same as the tokenizer used in Hunyuan-Large employing a vocabulary consisting of 128K tokens.

\subsection{Model Architecture}

Hunyuan-TurboS is a hybrid architecture that combines Transformer \citep{vaswani2017attention}, Mamba2 \citep{dao2024transformers}, and Feed-Forward Network (FFN) components. Designed in compliance with scaling laws and optimized for both training and inference efficiency, the model was scaled to 56B activated parameters and 560B total parameters. The architecture comprises 128 layers (where each Attention, FFN, and Mamba2 block counts as one layer). Each FFN layer adopts an MoE structure following similar strategies used in Hunyuan-Large \citep{sun2024hunyuan}, consisting of 1 shared expert and 32 specialized experts, with 1 shared + 2 specialized experts activated per forward pass. The Mamba2 layers employ a state-space model (SSM) architecture achieving linear sequence-length complexity (O(n)). We extensively explored various combinations of Attention (A), Mamba2 (M), and FFN (F) layers to optimize model performance while maintaining efficiency under fixed activated and total parameter budgets.
The ratio of Attention layers significantly impacts model performance, validation loss, key-value (KV) cache overhead, and inference efficiency.
To maximize training efficiency and model effectiveness, we maintained FFN layers at 50\% percent of the total, with 5.5\% Attention and 44.5\% Mamba2 layers.
The Attention layers utilize Grouped-Query Attention (GQA) to minimize KV cache memory usage. Additionally, QK normalization was implemented in the Attention blocks to enhance the training stability. 

For the specific architecture, We define the ``AMF'' block (Attention$\rightarrow$Mamba2$\rightarrow$FFN) emerged as an optimal atomic configuration, effectively balancing efficiency. We also adopt the ``MF'' block in our structure for further efficiency.
Hunyuan-TurboS employs an interleaved architecture of ``AMF'' and ``MF'' blocks. 

\textbf{Model Hyper-parameters.}
Hunyuan-TurboS uses a hidden dimension of 5,120, while each expert's intermediate dimension is set to 17,024.
In the Attention layers, we configure 64 Attention heads with 8 KV heads. For the Mamba2 blocks, we use 64 parallel heads with a SSM group size of 16. The chunk size in Mamba2 layer is 128.

\textbf{Training Hyper-parameters.}
Key hyper-parameters of Hunyuan-TurboS are listed in Table \ref{tab:overall_para}. We train the base model with a sequence length of $4,\!096$ tokens for a total of $16$ trillion tokens. The optimization uses AdamW ($\beta_1=0.9$, $\beta_2=0.95$) with weight decay $\lambda=0.1$. For Mixture of Experts training, we set the capacity factor $\gamma=1.5$ to ensure adequate expert coverage. 

\begin{table}[!hbtp]
    \centering
    \caption{Overview of the key hyper-parameters of Hunyuan-TurboS.}
        \begin{tabular}{l|c}
            \toprule
            \textbf{Configuration} & \textbf{Value} \\
            \midrule
            \# Layers & 128\\
            \# Attention Heads & 64\\
            \# Key/Value Heads & 8\\
            \# Mamba2 SSM Groups & 16\\
            \# Shared Experts of MoE Layers & 1\\
            \# Specialized Experts of MoE Layers & 32\\
            \# Activated Specialized Experts of MoE Layers & 2\\
            \# Trained Tokens & 16T\\
            Mamba2 d\_state size & 128 \\
            Mamba2 chunk size & 128\\
            Vocabulary Size & 128K\\
            Hidden Size & 5120\\
            \bottomrule
        \end{tabular}

    \label{tab:overall_para}
\end{table}

\subsection{Annealing}

Following the completion of our pre-training stage, we introduce an annealing stage designed to rapidly decay the learning rate and refine the capabilities of the base model. This stage begins from the terminal learning rate of pre-training and applies a fast cosine decay schedule down to a minimal learning rate of 5e-6. We configure a sequence length of 4,096 tokens and a batch size corresponding to approximately 9 million tokens. We allocate 300B tokens for the annealing stage. This setup facilitates efficient large-context training while maintaining computational tractability. To ensure comprehensive performance and robustness, the annealing stage is trained on a heterogeneous mixture of data, including high-quality pre-training data, code, mathematics, STEM-related corpora, instruction-following datasets (such as long-CoT data), and other synthetically generated samples. We conduct extensive data ablation and composition studies to optimize the data distribution, addressing model deficiencies and enhancing generalization.
Several key insights that emerged during this stage are as follows.

\begin{enumerate}[label=(\arabic*)]
    \item Retaining high-quality data from pre-training is of critical importance. Through extensive empirical studies, we carefully curated the composition and proportion of this subset, and found that such optimization leads to substantial improvements in both downstream task performance and generalization ability.
    \item A substantial portion of instruction-following data is advanced into the annealing stage. This strategy reduces the amount of instruction tuning required during the SFT stage and facilitates greater capacity for improvement in the subsequent RL stage.
    \item A moderate proportion of web-crawled content is retained, and target loss is selectively applied to QA-style data. This combination helps prevent the base model from drifting into generating test-like questions when it is supposed to continue open-ended text.
    \item Sequence truncation is carefully processed, particularly for long-CoT data. To avoid discarding critical answer content, sequences exceeding the context window are deferred to the long-context training phase.
\end{enumerate}

\subsection{Long-Context Extension}

In the final phase of pre-training, we employ a curriculum-based strategy to progressively expand the model's context window. The context is scaled from 4K tokens to 32K tokens, and ultimately to 256K tokens, as in Hunyuan-Large. This staged expansion is facilitated by NTK-aware \citep{peng2023ntk} positional encoding, incorporating scaling parameters $\alpha = 50$ for the 32K stage and $\alpha = 1000$ for the 256K stage. Throughout this stage, we maintain a constant learning rate of 5e-6 and a batch size of approximately 9 million tokens, consistent with the annealing stage. To balance capacity preservation with context extension, we adopt a cautious training strategy that leverages slightly increased token volumes, i.e., 30B for 32K and 20B for 256K, with a 3:1 ratio of short-context to long-context data. We also carefully curate a collection of long documents that span diverse domains and genres, which are relatively scarce in natural distributions but play a crucial role in enhancing the performance of downstream tasks.

\subsection{Evaluations on Pre-Trained Model}
\label{sec:pre_eval}

In this section, we evaluate the performance of Hunyuan-TurboS pre-trained model across a broad range of widely-used benchmarks, demonstrating its strong fundamental capabilities in diverse tasks.

\subsubsection{Benchmarks and Experimental Settings}

\textbf{Key Benchmarks.}
We evaluated Hunyuan-TurboS on a comprehensive set of widely-used benchmarks spanning multiple tasks, including commonsense reasoning, reading comprehension, question answering, mathematical problem solution, coding, and aggregated tasks, in both English and Chinese. We conduct extensive evaluations on the following benchmarks:

\begin{itemize}[leftmargin=10pt]
    \item \textbf{Aggregated knowledge:}
    We evaluate knowledge coverage using MMLU \citep{hendrycks2021measuring}, MMLU-Pro \citep{wang2024mmlu-pro}, MMLU-Redux \citep{gema2024we}, BBH \citep{suzgun2022challenging}, CMMLU \citep{li2023cmmlu}, C-Eval \citep{huang2024ceval}, CCPM \citep{li2021ccpm}, and SuperGPQA \citep{du2025supergpqa}.
    
    \item \textbf{Commonsense reasoning:}
    For commonsense understanding, we use HellaSwag \citep{zellers2019hellaswag}, WinoGrande \citep{sakaguchi2021winogrande}, and PIQA \citep{bisk2020piqa}.
    
    \item \textbf{Question answering \& reading comprehension:}
    We assess fundamental NLP abilities with DROP \citep{dua2019drop} and NaturalQuestions \citep{kwiatkowski2019natural}, while ARC-C \citep{clark2018think} and TriviaQA \citep{joshi2017triviaqa} evaluate science and world knowledge.

    \item \textbf{Mathematical reasoning:}
    Mathematical proficiency is tested via GSM8k \citep{cobbe2021training}, MATH \citep{hendrycks2021measuring}, CMATH \citep{wei2023cmath}, and MGSM \citep{shi2022language}.

    \item \textbf{Coding:}
    Coding ability is evaluated using EvalPlus \citep{chen2021evaluating}, MultiPL-E \citep{cassano2022multipl}, MBPP \citep{austin2021program}, CRUXEval \citep{gu2024cruxeval}, and LiveCodeBench \citep{jain2024livecodebench}.
    
\end{itemize}

\textbf{Evaluation Settings and Baselines.}
We adhere to standard evaluation protocols across benchmarks, including established metrics and shot configurations. Specifically, we employ zero-shot for TriviaQA, PIQA, C3, EvalPlus, MultiPL-E, LiveCodeBench, CRUXEval, and CCPM; 3-shot evaluation for BBH, MBPP, and DROP; 4-shot evaluation for GSM8K, MATH, and CMATH; 5-shot evaluation for MMLU, MMLU-Pro, MMLU-Redux, SuperGPQA, C-Eval, CMMLU, WinoGrande, and NaturalQuestions; 7-shot evaluation for CommonsenseQA; 8-shot evaluation for MGSM; 10-shot evaluation for HellaSwag; 25-shot evaluation for ARC-C. We compare Hunyuan-TurboS against state-of-the-art pre-trained base models of comparable parameter scales, including: Llama-4-Maverick \citep{Meta2025llama4}, DeepSeek-V3 \citep{liu2024deepseek}, and Qwen3-235B-A22B \citep{yang2025qwen3technicalreport}. Owing to the architectural advantage of Mamba2 and Attention hybrid design, our model achieves significantly lower downstream deployment costs. For fairness, we report the highest performance among publicly available results.

\begin{table}[!hbtp]
\scriptsize

	\centering
        \caption{Performance of pre-trained Hunyuan-TurboS and other open-source base models.}
	\resizebox{0.9\linewidth}{!}{
	\begin{tabular}{lcccc}
\toprule
\textbf{Model}       & \textbf{LLama-4-Maverick} & \textbf{DeepSeek-V3} & \textbf{Qwen3-235B-A22B} & \textbf{Hunyuan-TurboS} \\ \hline
Architecture             & MoE        & MoE           & MoE         & MoE           \\
\# Activated Params          & 17B          & 37B           & 22B         & {56B}           \\
\# Total Params              & 402B          & 671B          & 235B        & {560B}          \\
Context Length               & 256K         & 128K           & 128K     & 256K          \\ \hline
\textbf{English}                                                                      \\
MMLU                        & 85.16         & 87.19          & 87.81        & \textbf{87.94} \\
MMLU-Pro             & 63.91         & 59.84          & \textbf{68.18}           & 65.11          \\
MMLU-Redux            & 84.05         & 86.14          & \textbf{87.40}           & 87.11          \\
BBH                       & 83.62         & 86.22          & 88.87        & \textbf{89.76} \\
SuperGPQA                        & 40.85         & 41.53          & 44.06        & \textbf{54.63} \\
HellaSwag                     & -            & \textbf{88.9} &  -     & 87.1          \\
WinoGrande                  &  -        &  84.9         &  -       & \textbf{86.5} \\
PIQA                           & -            & 84.7          & -        & \textbf{86.7} \\
NaturalQuestions                & -            & 40.0          & -        & \textbf{45.6} \\
DROP                         & -         & \textbf{89.0}          & -        & 85.3 \\
ARC-C               & -         & 95.3          & -        &    \textbf{97.32}       \\
TriviaQA                       & -            & 82.9          & -        &  \textbf{92.22} \\ \hline
\textbf{Chinese}                                                                      \\
CMMLU                           & -            & 88.8          & -        & \textbf{90.57} \\
C-Eval                         & -            & \textbf{90.1}          & -        & 88.7  \\
CCPM                & - & \textbf{92.0} & - & 90.11 \\ \hline
\textbf{Math}                                                                         \\
MGSM                & 79.69 & 82.68 & \textbf{83.53} & 77.64 \\
GSM8K                       & 87.72         & 87.57          & \textbf{94.39}        & \textbf{94.39}  \\
MATH                         & -         & 61.6          & -        & \textbf{81.4} \\ %
CMATH                          & -            & 90.7          & -        & \textbf{91.83}  \\ \hline
\textbf{Code}                                                                         \\
EvalPlus                   & 68.23             & 63.75           & 77.60        & \textbf{78.99} \\
MultiPL-E          & 57.28 & 62.26 & 65.94 & \textbf{67.13} \\
MBPP                        & 75.40         & 74.20          & 81.40        & \textbf{85.45} \\ 
CRUX-I             & - & 67.3 & - & \textbf{68.88} \\
CRUX-O              & 77.0 & 69.9 & \textbf{79.0} & 75.88 \\
LiveCodeBench              & - & 19.4 & - & \textbf{23.49} \\ 

\bottomrule
\end{tabular}
	}
	\label{tab:main_pre_train}
\end{table}

\subsubsection{Model Performance of Pre-Training} 

As illustrated in Table \ref{tab:main_pre_train}, the results demonstrate the robust capabilities of Hunyuan-TurboS compared to other state-of-the-art models. As the largest model in this comparison with 56B activated parameters and 560B total parameters, Hunyuan-TurboS delivers leading performance across multiple domains while maintaining efficient inference capabilities through optimized hybrid Mamba2, Attention, and MoE architecture.

In English tasks, Hunyuan-TurboS achieves first-tier results, notably attaining 54.63 on SuperGPQA (exceeding all competitors by over 10\%) and 89.76 on BBH. While slightly trailing Qwen3-235B-A22B in MMLU-Redux (87.11 vs 87.40), our model demonstrates particularly strong reasoning capabilities with 92.22 on TriviaQA. Besides, Hunyuan-TurboS shows balanced strengths across various domains, achieving nearly state-of-the-art performance in WinoGrande (86.5) and PIQA (86.7). For Chinese capabilities, Hunyuan-TurboS reaches new height with 90.57 on CMMLU while maintaining competitive performance on C-Eval (88.7). Mathematical reasoning showcases Hunyuan-TurboS's most dramatic improvements, achieving 81.4 on MATH while maintaining parity with Qwen3-235B-A22B on GSM8K (both 94.39). Coding tasks emerges as another standout domain, where our model achieves the overall highest score compared to other models.

%% file: content/posttrain.tex
\newpage

\begin{figure}[hbp]
    \centering
    \includegraphics[width=0.99\textwidth, trim=0 60 0 10,clip]{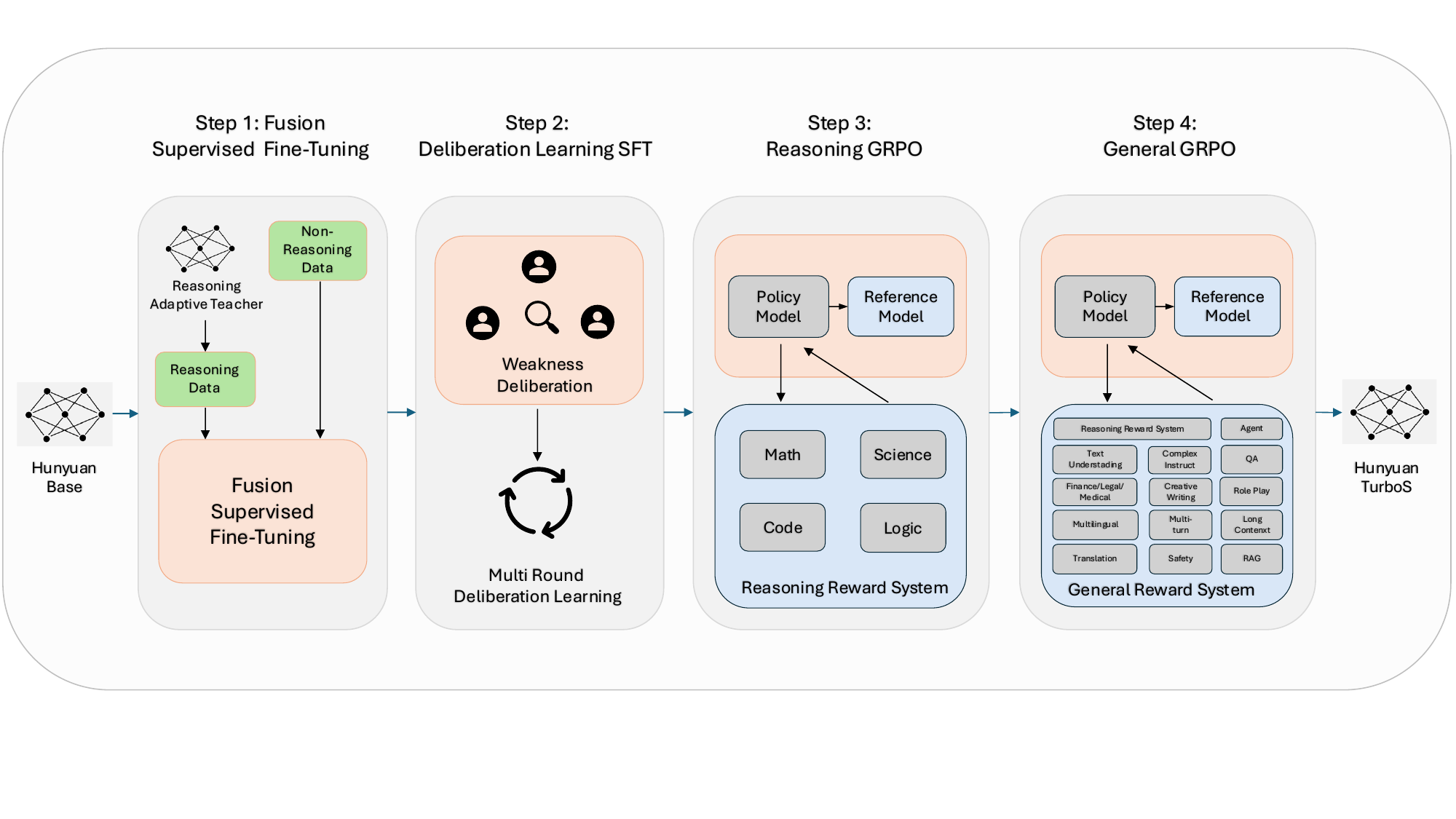}
    \caption{A diagram illustrating the four steps of Hunyuan-TurboS post-training.}
    \label{fig:post_train}
\end{figure}

\section{Post-training}

\label{sec:post}

As shown in the Figure~\ref{fig:post_train}, our approach to post-train Hunyuan-TurboS encompasses four critical modules:

\begin{enumerate}[label=(\arabic*)]

    \item \textbf{Supervised Fine-Tuning}: We categorize instruction data collection by comprehensively delineated topics across diverse domains. This methodical approach facilitates specialized data curation while implementing multi-dimensional metrics for rigorous filtering and quality assurance, establishing a robust foundation for the model's fundamental capabilities.
    
    \item \textbf{Adaptive Long-short Chain-of-Thought Fusion}: We propose a novel method to enable the model to autonomously select optimal reasoning strategies and efficiently allocate computational resources based on task requirements. This approach also enhances the readability of responses through lossless compression and reformatting of lengthy chains of thoughts.
    
    \item \textbf{Multi-round Deliberation Learning}: The SFT model will compete with other cutting-edge Hunyuan models across an extensive instruction collection, with a strategic focus on identifying and addressing capability gaps between Hunyuan-TurboS and other top-tier Hunyuan models.
    
    \item \textbf{Two-stage Large-scale Reinforcement Learning}: Our reinforcement learning begins with STEM-focused GRPO training to enhance reasoning capabilities, followed by general domain GRPO to improve instruction-following proficiency across all areas.
    
\end{enumerate}

\subsection{Supervised Fine-tuning}

This section details the supervised fine-tuning (SFT) phase of Hunyuan-TurboS. Quality and diversity of SFT data are critical for LLM performance across diverse tasks. We categorize SFT data into fine-grained topics, curating high-quality samples for each and integrating them into a unified dataset. The data construction for each topic is detailed below.

\begin{enumerate}[label=(\arabic*)]
    \item \textbf{Math:} 
    We collect math problems from diverse educational sources (textbooks, exams, competitions), spanning difficulties from elementary to academic competitions. Generative reward models and verifiers ensure CoT quality through evaluation and iterative refinement.

    \item \textbf{Coding:} 
    A pipeline creates high-quality instruction data from source code. Code snippets from open-source repositories (e.g., GitHub) are transformed into instructional pairs~\citep{magicoder}, categorized and optimized for diverse task types, languages, and knowledge. Data quality is ensured by multi-stage filtering with critic models and sandbox execution.

    \item \textbf{Logic:} 
    We extract data from public/licensed sources and use an automated synthesis pipeline (akin to ZebraLogic~\citep{zebralogic}) to scale volume. Data is categorized by question type and difficulty. Quality is ensured by tiered validation: models for standard cases, human experts for complex ones, balancing accuracy and efficiency.

    \item \textbf{Science:} 
    We collected diverse data (physics, chemistry, biology) from middle school to graduate level. LLMs mark difficulty/quality; difficult questions (e.g., Olympiads, university-level) are selected for post-training. An LLM-based Verifier with CoT checks if generated answers match references, handling complex checks like unit conversion, approximations, and equivalent forms (e.g., chemical names, equations).

    \item \textbf{Language-Centric Tasks:}
    Focuses on language understanding, translation, and generation. Strict filtering and rewriting ensure data quality; model verification removes meaningless/ambiguous instructions. Generative reward models score responses via joint evaluation of paired responses, mitigating reward hacking. Expert rewriting and iterative refinement further optimize outputs.

    \item \textbf{Creative Writing:}
    Multi-dimensional labels (genre, style, etc.) ensure instruction set richness and diversity. A discriminant RM filters samples with low scores/high variance to retain valuable instructions. Expert rewriting and model self-refinement build high-quality responses.

    \item \textbf{English and Multilingual:}
    Diverse instruction data are created through document augmentation, instruction evolution~\citep{xu2024wizardlm}, and back-translation~\citep{li2023self}. A well-trained team of human experts will annotate high-quality responses.
    
    \item \textbf{Complex Instruction:} 
    For complex instructions, we vary constraint numbers/types for difficulty, with rule-based filters checking satisfaction. For long-context, questions require integrating multiple context segments. For agent, diverse scenarios are created by combining tool usage, decision types, and multi-turn dynamics.

    \item \textbf{Role Play:} Diverse character profiles are created from typical personality traits. Conversational data is generated via prompt engineering, with responses assessed for instruction comprehension, trait representation, and emotional empathy.
    
    \item \textbf{Knowledge QA:} To reduce hallucinations, we optimize performance on knowledge-intensive tasks. For general knowledge, cross-validation and critic models filter errors and select best answers ~\citep{ke2024critiquellminformativecritiquegeneration,wang2024halujcritiquebasedhallucinationjudge}.

    \item \textbf{Multi-turn:} 
    Multi-turn dialogues are divided into 6 categories. For each, diverse data is constructed via open-source collection, procurement, synthesis, instruction evolution, and pseudo-multi-turn synthesis. 

    \item \textbf{Finance/Legal/Medical:} Experts in legal, financial, and medical fields label data for high-quality training sets.
    
    \item \textbf{Safety:} Nearly 1,000 safety categories measure diversity and coverage to address vulnerabilities. A classification model continuously identifies risky data. A red team conducts adversarial testing to uncover security flaws.
\end{enumerate}

Based on above topics, we create an SFT dataset of 3 million samples (reasoning and non-reasoning). Complex reasoning tasks requiring longer CoT undergo additional processing:
\begin{enumerate}[label=(\arabic*)]
\item \textbf{Reasoning Data.} Math, coding, science and logic are categorized as reasoning data. While models like DeepSeek-R1~\citep{deepseek_r1} and OpenAI-o1~\citep{o1} improve on such tasks, they can overthink simple questions. We create adaptive long-short CoT responses using an internal teacher model (Section~\ref{subsec:long_shot_fusion}) to address this. Unlike prior work~\citep{deepseek_v3, yang2025qwen3} using system prompts, our reasoning data is uniformly structured as \(<\)problem, response\(>\), with responses potentially including detailed reasoning. This allows the model to dynamically activate thinking mode based on problem difficulty.

\item \textbf{Non-Reasoning Data.} For non-reasoning data, original responses from diverse topics are used directly in SFT.
\end{enumerate}

\subsection{Adaptive Long-short Chain-of-Thought Fusion}
\label{subsec:long_shot_fusion}

The short chain-of-thought mode excels at quick, heuristic decision-making and is suitable for solving simple problems. The long chain-of-thought mode relies on deep reasoning to make more accurate judgments and reduce biases, making it appropriate for complex problems. If an LLM operates solely in the short CoT mode, it struggles with complex and difficult problems. Conversely, if it operates only in the long CoT mode, many everyday problems requiring quick responses would consume excessive inference resources. We propose an adaptive long-short CoT fusion method that creatively integrates these two reasoning modes into a single model, allowing the LLM to autonomously decide whether to use long or short CoT and determine the depth of reasoning based on the complexity of the problem. Previous studies~\citep{deepseek_r1,o1} have revealed that long CoT is particularly effective in reasoning fields such as mathematics. Therefore, we apply adaptive long-short CoT fusion for reasoning data (e.g., mathematics, STEM), while primarily utilizing the short CoT mode for non-reasoning data. We trained an adaptive long-short CoT fusion teacher model to generate training data for the first SFT stage of Hunyuan-TurboS. The training of this teacher model includes two stages: supervised fine-tuning and reinforcement learning.

\subsubsection{Adaptive Long-Short CoT SFT Training}

First, we train Hunyuan-Base to obtain a short CoT model using reasoning data. Then, we use this model to infer answers on all reasoning data and perform consistency checks. For one data sample, if the short CoT model gives a correct answer, it will be directly added as a training sample. If the short CoT model gets it wrong on the first attempt, we will feed the question and the short CoT's incorrect response into Hunyuan-T1 to continue generating the subsequent reasoning process and answer, and then convert this extended reasoning process and answer into the short CoT response style. We will repeatedly apply this Hunyuan-T1 generation process until a correct answer is obtained for the given question. Then, we will concatenate all failed attempts along with the correct response to serve as the training response for our adaptive long-short fusion teacher model.Finally, we use the adaptive short-long CoT data generated through the process described above to train Hunyuan-Base, thereby obtaining an adaptive SFT model.

\subsubsection{Reinforcement Learning for Adaptive Long-Short CoT}

Our long-short adaptive reward framework enables LLMs to choose appropriate thinking modes based on problem difficulty:

\begin{enumerate}[label=(\arabic*)]

    \item \textbf{Difficulty-Adaptive Reward:} During GRPO sampling, we generate responses with varying depths of reasoning for each prompt. An online rejection sampling mechanism evaluates prompt difficulty and selects the appropriate mode—assigning long CoT to complex problems and short CoT to simpler ones.

    \item \textbf{Long CoT Compression Reward:} For long reasoning chains, we apply a length penalty during the calculation of the reward. When multiple reasoning paths achieve equal correctness, shorter traces receive higher rewards, minimizing redundancy while maintaining accuracy.

\end{enumerate}

The proposed reinforce framework generates significantly adaptive CoTs, achieving a balance between reasoning quality and computational efficiency.

\subsection{Deliberation Learning}

To enhance the capabilities of our Hunyuan-TurboS after its initial pre-training and foundational SFT, and inspired by~\cite{luo2024wizardarena}, we propose a new iterative refinement strategy of human-LLM collaboration grounded in the principles of Deliberation Learning. This approach leverages a data flywheel, where models progressively improve by competing with each other, with weakness profiles identified by powerful LLM-based judges and human experts to inform subsequent SFT iterations.

\subsubsection{Training Powerful Judge LLMs to Simulate Human Annotators}

To effectively simulate human annotators and mitigate individual LLM biases, we developed and trained a panel of Judge Models based on Hunyuan-TurboS. Instead of relying on a single holistic score, model responses were assessed across multiple predefined dimensions: accuracy, helpfulness, harmlessness, coherence, conciseness, and adherence to instructions. For each pairwise comparison, each judge provided scores along these dimensions, often supplementing them with textual rationales. A consensus mechanism, such as majority voting or a weighted scoring system, aggregated these multi-dimensional judgments. This multi-judge, multi-dimensional approach was designed to bolster the robustness of our evaluations and more closely align automated judgments with human preferences. Furthermore, the judging protocol underwent regular calibration against evaluations from human experts to ensure its continuous refinement.

\subsubsection{Build a Data Flywheel to Post-train Hunyuan-TurboS}
\label{subsec:data_flywheel_hunyuan}

Our core innovation lies in an iterative improvement cycle that continuously enhances Hunyuan-TurboS' capabilities through competitive evaluation and targeted Supervised Fine-Tuning (SFT). This cyclical process encompasses three key phases:

\begin{enumerate}[label=(\arabic*)]

    \item \textbf{Judging:} We initialize a competitive environment with our Hunyuan-TurboS SFT model and our cutting-edge Hunyuan models, such as Hunyuan Large, Hunyuan Turbo, and Hunyuan T1.  In each pair-wise comparison, all models generate responses to identical prompts from our curated training split. These responses are then meticulously assessed by our multi-LLM judge ensemble.

    \item \textbf{Weakness Deliberation:} We identify model weaknesses using human expert and LLM oversight. While automated metrics offer initial indicators, domain experts review intricate comparison outcomes and nuanced model failures missed by automated systems. Experts compare Hunyuan-TurboS with competitors, using deep contextual understanding to find capability gaps across diverse domains, tasks, and challenging prompts. This expert analysis, often with collaborative adjudication, creates a detailed "weakness profile" informing data selection and augmentation.

    \item \textbf{Iterative SFT:} Guided by this profile, we develop tailored training batches for identified deficiencies frequently incorporating "loss data". These training batches will be carefully annotated by human experts with high-quality outputs, which will be added incrementally to the training process. Our approach uses curriculum learning, progressively increasing task complexity and skill subtlety as the model shows mastery. After each SFT iteration, the updated model is re-evaluated, closing the loop and perpetuating the data flywheel. This cyclical refinement yields incremental enhancements for identified weaknesses. Our methodology aims to preserve strengths while addressing deficiencies, fostering robust and well-rounded performance.

\end{enumerate}

\subsection{General Reward System}

To facilitate effective reinforcement learning, we design a general reward system organized around three key components: a \emph{Generative Reward Model} with reference answers that covers most scenarios, an \emph{Answer Consistency Model} for tasks such as mathematics where ground-truth answers exist, a code \emph{Sandbox} that executes unit tests for programming problems. Finally, a \emph{Reward Aggregation} module incorporates domain-specific rules to produce a unified score. Overall, the system spans 16 sub-topics and more than 30 scoring services, each of which relies on specialized models or rule-based heuristics tailored to its specific evaluation scenario.

\textbf{Generative Reward Model with Reference Answer.} 
Following \citep{zhang2024generative, mahan2024generative, xu2025unifiedpairwiseframeworkrlhf}, we employ a generative reward model (GRM) that compares candidate answers against a \emph{reference answer}. For tasks with a single deterministic solution, such as closed-book factual QA, the reference is the ground-truth answer. For open-ended tasks (e.g., creative writing or open-domain dialogue) we still provide a carefully curated reference; however, the GRM treats it only as a semantic anchor rather than expecting an exact match.  

We train the GRM using a \emph{pairwise} preference scheme: given two candidate answers to the same prompt, it predicts which one is better. To construct the training corpus, we sample diverse responses from multiple Hunyuan-TurboS checkpoints, ensuring broad coverage. Human annotators then label these pairs, achieving inter-annotator agreement above 93\%. In total, we collect about 200K high-confidence annotations. To mitigate the well-known \emph{position bias}, we swap the left/right order of the candidates to augment the training data.

The GRM can optionally ingest Chain-of-Thought (CoT) reasoning traces, yielding a variant we call \emph{GRM-CoT}. Supplying rationales markedly improves judgment accuracy on multi-step reasoning tasks, albeit at increased computational cost. We also employ a \emph{critic model} capable of invoking external tools for verification. The critic reports explicit length statistics and checks outputs against task-specific constraints. Leveraging the GRM-CoT approach together with carefully crafted prompts, the critic assesses answer correctness. 

\textbf{Answer Consistency Model.}
This lightweight classifier verifies if the generated final answer matches the reference, providing a binary reward (1 for a match, 0 otherwise). It is designed to normalize superficial differences, such as variations in whitespace, unit formats, or synonyms, to minimize false negatives. Due to its speed and robustness, this model serves as both a hard filter during data curation and a complementary reward signal during RL training.

\textbf{Sandbox.}
We have built a multilingual code sandbox that supports 36 programming languages. These include, but are not limited to, Python, C, C++, Java, Go, JavaScript, C\#, CoffeeScript, Common Lisp, Dart, Elixir, Emacs Lisp, Erlang, F\#, Fortran, Groovy, Haskell, Julia, Kotlin, Lua, Pascal, Perl, PHP, PowerShell, Racket, R, Ruby, Rust, Scala, Scheme, Shell, Swift, Tcl, TypeScript, VimScript, and Visual Basic. The code sandbox is deployed on a distributed CPU cluster capable of handling over 1000 concurrent executions. Strict security measures such as file and network isolation are in place to prevent the execution of harmful code.

To address imbalances and insufficient diversity in the post-training data such as skewed distributions and a lack of challenging cases, we have developed a multi-level knowledge classification framework that covers over 400 categories and 13K tags. This system ensures diverse prompt generation through data distribution adjustments and incorporates a difficulty grading mechanism based on the number and complexity of instructions. 

By leveraging high-quality open-source seed code fragments and pre-training data, we synthesize executable code complete with unit tests. Our approach considers boundary cases in inputs to yield verified outputs via sandbox execution. In total, over 800K executable, unit-tested data samples have been created, with mainstream programming languages (Python, C, C++, Java, Go, and JavaScript) contributing 100K samples each and long-tail languages contributing 10K samples each.

\textbf{Reward Aggregation.}
A single prompt may trigger multiple scoring services. The outputs from these services are  applied with domain-specific custom rules and then combined by scoring fusion. We aggregate complementary scores, for example, by merging reward scores with repetition penalties, or by combining scores from creative evaluation models with those from critic models. During RL training, directly using raw reward values can lead to instability in advantage estimation. We therefore introduce group-level normalization. For each prompt, rewards within a group of generated responses are rescaled such that poorly performing answers receive negative advantages and well-performing answers receive positive advantages. This ensures more stable policy updates during RL training.

\subsection{RL Training}

To align and optimize our language models for diverse, domain-specific requirements, we adopt an incremental, domain-focused reinforcement learning (RL) pipeline based on the Generative Reward Preference Optimization (GRPO) framework \citep{shao2024deepseekmath}. Our GRPO training involves a two-stage strategy, detailed below, alongside key implementation choices. 

\subsubsection{Two-Stage GRPO Training Strategy}

Our GRPO training incrementally integrates knowledge and optimization objectives from multiple domains. For each domain, we curate specific datasets and define training recipes. The reward signals for these domains are derived from our General Reward System, and the resulting optimization tasks are merged into the main GRPO curriculum.

In Stage I, we enhance reasoning abilities (e.g., in domains such as Logic, Coding, Mathematics, and Science). In Stage II, we boost general performance (e.g., in domains such as Text Understanding, Translation, Long Context processing, and Creative Writing). We present the details of the two-stage GRPO with corresponding topic domains.

\textbf{Stage I: Reasoning GRPO.}
This stage targets Logic, Coding, Mathematics, and Science domains. 300K training data are mixed with a ratio of Code : Mathematics : Logic\&Science = 2 : 2 : 1. Given that the Supervised Fine-Tuned (SFT) backbone model already demonstrates strong performance on these tasks and exhibits low output entropy, we apply a relatively small Kullback-Leibler (KL) divergence constraint to encourage broader exploration during this stage.

Each of the following reasoning domains is backed by its own reward service, dedicated preference data and evaluation rules, as well as domain-specific prompt construction.

\begin{enumerate}[label=(\arabic*)]

\item \textbf{Logic.}
Prompts exhibiting instability, identified through high variance in accuracy across multiple generated samples, are targeted. A curated set of short-answer, validated items is used to strengthen the model's logical robustness.

\item \textbf{Coding.}
Challenging programming prompts are selected and paired with automated evaluation systems. Rejection sampling and difficulty-aware filtering techniques are employed to curate high-quality data for RL.

\item \textbf{Mathematics.}
We collect challenging mathematical problems, excluding proof-based questions, multiple-choice, and true/false formats. Edge cases are converted into more tractable forms. Model-based rejection sampling is used to refine this corpus.

\item \textbf{Science.}
A generative verifier is trained to assess the consistency of answers against reference solutions. This model is capable of handling multi-step reasoning, unit conversions, approximations, formula equivalence, and chemical notations, thereby generating reward signals used for both rejection sampling and RL.

\end{enumerate}

\textbf{Stage II: General GRPO.}
In this stage, optimization is extended to general tasks, with a focus on balancing performance across various domains. We continue to include 10\% reasoning data from Stage I in the training mix. Hyperparameters from Stage I (e.g., clipping range, learning rate) are largely retained, but the KL divergence penalty coefficient is increased to mitigate issues like catastrophic forgetting or performance collapse during training on 160k general RL instructions.

To ensure balanced general capabilities, every subsequent track is supported by a dedicated reward service, corresponding preference datasets and rules, and specially designed  data construction pipelines.

\begin{enumerate}[label=(\arabic*)]

\item \textbf{Text Understanding.}
Two reward models are utilized: a consistency model for objective Q\&A and a comparative GRM for subjective or open-ended tasks.

\item \textbf{Translation.}
Domain experts annotate parallel corpora; GRMs trained on these annotations provide faithful reward signals.

\item \textbf{Long Context.}
For long-context tasks, an additional hallucination-focused critic model and online RL further enhance stability.

\item \textbf{Creative Writing.}
For creative writing, paired GRMs based on relative preference judgments are used to mitigate reward hacking. During RL, creative rewards are blended with automated checks for instruction adherence, balancing creativity, fluency, and compliance.

\item \textbf{Agents.}
Action-level rule-based rewards and extensive step-wise data markedly improve multi-step reasoning capabilities.

\item \textbf{Multi-Turn Dialogue.}  
We refine dialogue-specific critic models and general rewards, mining unstable conversations.

\item \textbf{Complex Instructions.}
We utilize constraint-extraction and satisfaction tools, complemented by general critic and reward models.

\item \textbf{Role-Playing.}
We evaluate instruction comprehension, character consistency, and empathy, then generate corresponding data using generalized critic and reward models.

\item \textbf{Safety.}
For safety alignment, safe response pairs are identified using classifiers and refusal heuristics, then incorporated into preference datasets.

\item \textbf{Knowledge QA.}
For knowledge question-answering, rewards from hallucination detection models (both with and without access to references) and user-experience-focused models are jointly optimized to reduce hallucination risk.

\item \textbf{Multilingual.}
For multilingual capabilities, SFT answers are sampled and scored using GRMs. A prompt is kept if it has high diversity, high answer variance and high quality.

\item \textbf{Finance, Legal, and Medical Domains.}
For specialized domains like Finance, Legal, and Medical, consistency-based rewards identify unstable items from professional exams, which then supply preference data for targeted domain-specific improvements.

\end{enumerate}

\subsubsection{More Details on GRPO Implementation}

Practical experience indicates that several engineering choices are decisive for achieving GRPO training that is both \emph{stable} and \emph{sample-efficient}. Following \citep{liu2024deepseek, yu2503dapo}, we mention some additional details applied, unless otherwise noted, across all domains and training stages. 

\textbf{GRPO Loss.}
We reformulate the GRPO loss at the token level, which markedly improves KL-stability during training. An approximate K3 KL loss term is clipped to the range [0, 10] to prevent issues arising from extreme log-probability spikes and potential divergence or model collapse.
Best-of-N (BON) loss is applied selectively only to samples exhibiting positive advantage. This strategy helps preserve policy entropy, prevent premature convergence, and encourage continued exploration.

\textbf{Prompt Filtering.}
We filter prompts by excluding \emph{extreme} cases where the model consistently succeeds or fails. Conversely, we retain \emph{unstable} prompts, for which the model's sampled outputs show substantial disagreement, as these provide ideal adversarial examples for RL.

\textbf{Sampling.}
The sampling temperature for generating responses during RL is set to 1.0. Experiments with lower temperatures indicated that they led to rapid entropy decay, which hindered exploration and ultimately capped performance improvements. As model capabilities evolve during training, prompts that were once difficult may become trivial. Inspired by \citep{yu2503dapo}, we implement a dynamic sampling strategy. Samples yielding zero advantage are dropped during batch formation, which helps accelerate convergence and improve overall training stability.

\textbf{Group Reward Adjustment.}
Standard relative normalization can sometimes assign unintended positive advantages to poor-quality answers. Our \textit{group reward adjustment} scheme addresses this by rescaling rewards within each prompt's response group, ensuring that undesirable responses receive negative advantages and desirable ones receive positive advantages, thereby promoting stable policy updates.

%% file: content/experiments.tex
\section{Hunyuan-TurboS Evaluation}
\label{sec:experiment}

\subsection{LMSYS Chatbot Arena}
    
In this section, we report the performance of our model Hunyuan-TurboS-20250416 on LMSYS Chatbot Arena~\citep{chiang2024chatbot}. LMSYS Chatbot Arena uses blind side-by-side evaluations by human raters with less biases and more objective evaluation of the chatbots' capabilities. We report the overall rank in Table \ref{tab:hunyuan_TurboS_LMSYS} and Appendix \ref{sec:appendix}, and the detailed results are as following:

\begin{enumerate}[label=(\arabic*)]
    \item Hunyuan-TurboS-20250416 obtains an overall score of \textbf{1356}, is among the \textbf{top 7} best models in total 239 models, and outperforming leading reasoning models like o4-mini-2025-04-16.

    \item Hunyuan-TurboS-20250416 achieves \textbf{top 1} in \textbf{Chinese}, \textbf{French}, \textbf{Spanish}, and \textbf{top 2} in \textbf{Korean} languages. This shows the comprehensiveness and advancement of our model in terms of multilingualism and internationalization.

    \item Hunyuan-TurboS-20250416 achieves \textbf{top 5} in various main Arena tasks, such as \textbf{Hard Prompts}, \textbf{Creative Writing}, \textbf{Multi-Turn}, and \textbf{Longer Queries}.

\end{enumerate}

\begin{table}[htbp]
\centering
\small

\begin{adjustbox}{max width=0.98\textwidth}
\renewcommand{\arraystretch}{1.2} %

\begin{tabular}{clcccccc}
\hline
Rank & Model & Arena Score & 95\% CI & Votes & Organization  \\
\hline
1 & Gemini-2.5-Pro-Preview-05-06 & 1446 & +8/-7 & 5696 & Google  \\
2 & o3-2025-04-16 & 1409 & +5/-8 & 7621 & OpenAI  \\
2 & ChatGPT-4o-latest (2025-03-26) & 1405 & +6/-6 & 10284 & OpenAI \\
2 & Grok-3-Preview-02-24 & 1399 & +4/-5 & 14845 & xAI \\
4 & GPT-4.5-Preview & 1395 & +3/-5 & 15275 & OpenAI \\
4 & Gemini-2.5-Flash-Preview-04-17 & 1389 & +8/-6 & 6622 & Google  \\
7 & DeepSeek-V3-0324 & 1369 & +5/-6 & 9404 & DeepSeek \\
7 & GPT-4.1-2025-04-14 & 1365 & +7/-7 & 5778 & OpenAI \\
7 & \textcolor{blue}{Hunyuan-TurboS-20250416} & \textcolor{blue}{1356} & \textcolor{blue}{+7/-7} & \textcolor{blue}{4863} & \textcolor{blue}{Tencent} \\
8 & DeepSeek-R1 & 1355 & +3/-4 & 19066 & DeepSeek \\
9 & Gemini-2.0-Flash-001 & 1352 & +5/-4 & 24922 & Google \\
9 & o4-mini-2025-04-16 & 1345 & +7/-9 & 5763 & OpenAI  \\
10 & o1-2024-12-17 & 1347 & +3/-3 & 29038 & OpenAI \\
11 & Mistral Medium 3 & 1339 & +10/-12 & 2838 & Mistral \\
12 & Qwen3-235B-A22B & 1339 & +9/-8 & 4457 & Alibaba \\
13 & Gemma-3-27B-it & 1338 & +5/-5 & 12721 & Google \\
13 & Qwen2.5-Max & 1338 & +3/-4 & 23176 & Alibaba \\
14 & o1-preview & 1332 & +3/-3 & 33173 & OpenAI \\
16 & Qwen3-32B & 1325 & +8/-6 & 3662 & Alibaba \\
18 & GPT-4.1-mini-2025-04-14 & 1320 & +7/-8 & 5616 & OpenAI \\
18 & Gemma-3-12B-it & 1318 & +9/-9 & 3593 & Google \\
19 & o3-mini-high & 1321 & +4/-5 & 19405 & OpenAI \\
19 & DeepSeek-V3 & 1315 & +4/-3 & 22834 & DeepSeek \\
21 & QwQ-32B & 1310 & +5/-4 & 9941 & Alibaba \\
21 & Gemini-2.0-Flash-Lite & 1310 & +3/-4 & 25654 & Google \\
21 & GLM-4-Plus-0111 & 1308 & +8/-5 & 6025 & Zhipu \\
21 & Qwen-Plus-0125 & 1307 & +7/-7 & 6058 & Alibaba \\

\hline
\end{tabular}
\end{adjustbox}
\caption{Leaderboard of LMSYS Chatbot Arena by May 18, 2025. More details are in \textbf{Appendix \ref{sec:appendix}.}}
\label{tab:hunyuan_TurboS_LMSYS}

\end{table}

\subsection{Auto Evaluation Results}

\begin{table*}
\label{tab:hunyuan_TurboS_benchmarks}
\scriptsize

\renewcommand\tabcolsep{5pt}

\renewcommand{\arraystretch}{1.2} %

\centering
\begin{tabular}{clccccccc cccc}
\hline

&  &\textbf{GPT4.5} &\textbf{Claude3.7} &\textbf{DeepSeekV3}  
&\textbf{DeepseekV3} &\textbf{Qwen2.5} &\textbf{Doubao1.5} &\textbf{Hunyuan-TurboS} \\

&  &&\textbf{Sonnet} & \textbf{0324} & &\textbf{max} & \textbf{Pro-32k}
&\textbf{-20250416} & \\

\hline

\hline
& GSM8k  & 91.9 & 86.6 & 93.3 & 91.6 & 92.3 & \textbf{95.6} & \underline{94.4}\\
 & MATH  & 86.2 & 56.8 & \underline{89.1} & 83.2 & 78.9 & {88.6} & \textbf{90.0}\\
${Math}$ & AIME2024  & 36.7 & 23.3 & \textbf{59.4} & 39.2 & 27.6 & {33.3} & \underline{56.7}\\
 & AIME2025  & 30.0 & 10.0 & \textbf{50.0} & 16.7 & 26.7 & {20.0} & \underline{40.0}\\
& OlympiadBench  & 67.6 & 54.6 & \textbf{79.5} & 55.9 & 53.4 & {59.8} & \underline{76.1}\\
\hline
& BBH  & 76.3 & {84.6} & \underline{90.9} & 90.4 & 89.6 & \textbf{91.6} & {90.8}\\
${Reasoning}$ & DROP & 83.3 & {85.9} & 91.5 & \underline{91.6} & 88.2 & \textbf{93.0} & 89.8\\
  & Zebra-logic & 53.7 & 47.4 & \textbf{84.7} & 40.6 & 27.7 & 36.0 & \underline{81.7}\\
\hline
$Code$  & HumanEval  & \underline{93.0} & \textbf{95.0} & \textbf{95.0} & 90.0 & \underline{93.0} & 91.0 & 89.0\\
$Tasks$ & LiveCodeBench  & 43.4 & {44.1} & 49.2 & 37.6 & 38.7 & 28.9 & 43.0\\
\hline
& MMLU  & 87.6 & 85.2 & 86.7 & \underline{88.5} & 86.4 & \textbf{88.6} & 85.8\\
$Knowledge$ & MMLU-pro & \underline{81.0} & 57.0 & \textbf{81.2} & 75.9 & 76.1 & 80.1 & 78.0\\
$Tasks$  & GPQA-diamond  & \underline{68.2} & {68.0} & \textbf{68.4} & 59.1 & 60.1 & 61.5 & 62.1\\
& Chinese-SimpleQA & \underline{72.5} & {59.0} & \textbf{73.6} & 64.8 & 69.0 & 63.6 & 69.6\\
\hline
$Chinese$  & C-Eval & 78.4 & 55.4 & 88.2 & 86.5 & \underline{89.4} & \textbf{91.8} & 88.1\\
$Tasks$ & CMMLU  & {82.7} & 77.9 & 88.6 & 85.5 & \underline{90.2} & \textbf{90.9} & 89.4\\
\hline
& LiveBench  & \textbf{70.0} & 62.0 & \textbf{70.0} & 61.0 & 62.2 & 60.3 & \underline{67.0}\\
& Arena-Hard  & \underline{92.1} & 91.5 & \textbf{94.6} & 86.2 & 91.4 & 81.2 & 91.9\\
$Alignment$ & AlignmentBench & \textbf{8.8} & 8.4 & \underline{8.7} & 8.4 & 8.4 & 8.5 &\textbf{8.8}\\
& MTBench  & \underline{9.2} & \underline{9.2} & 9.1 & 9.0 & 9.0 & 8.8 & \textbf{9.3}\\
& AlpacaEval  & 64.2 & 57.4 & \textbf{82.9} & 64.9 & 56.2 & 39.5 & \underline{76.0}\\
\hline
$Instruction$ & IF-Eval & {88.9} & \textbf{90.8} & 82.6 & 84.5 & 85.8 & \underline{89.5} & 87.6\\
$Following$  & SysBench & \textbf{82.9} & {80.4} & \underline{82.7} & 79.2 & 80.6 & 67.6 & 81.8\\

\hline
& AVG.  & 75.0 & 69.4 & \textbf{80.2} & 71.8 & 71.6 & 70.7 & \underline{77.9}\\
\hline

\end{tabular}
\caption{Comparison among Hunyuan-TurboS-20250416 and other AI models. The highest and second-best scores are shown in bold and underlined, respectively.}
\label{tab:model_comparison}
\end{table*}

\subsubsection{Evaluation Benchmarks}
In Table \ref{tab:model_comparison}, we show the performance of our model in key benchmarks, compared against several industry-leading models including open-source and closed-source. For closed-source models, evaluations are performed through their respective APIs. We conduct comprehensive evaluations of our model and focus on performance in knowledge, reasoning, mathematics, coding, Alignment Task, and instruction following. Hunyuan-TurboS achieves state-of-the-art performance on these dimensions and is comparable to leading models in the industry. 
\begin{itemize}
    \item \textbf{Mathematics}: For mathematics skill, we utilize math benchmarks including MATH(5-shot), GSM8k(4-shot), and high-level competitions including AIME(2024, 2025), and OlympiadBench.
    \item \textbf{Reasoning}: For logical reasoning skills, we employ high-level benchmarks including BBH(3-shot), DROP(3-shot), and ZebraLogic.
    \item \textbf{Coding}: To test the model's ability in coding tasks, we use livecodeBench (2408-2411)  and HumanEval. 
    \item \textbf{Knowledge \& Chinese Tasks}: We utilize benchmarks including MMLU(5-shot), MMLU-Pro(5-shot), GPQA-Diamond, C-Eval(5-shot), Chinese SimpleQA, and CMMLU(5-shot). 
    \item \textbf{Alignment Tasks}: To assess alignment with human preferences on
general topics, we utilize AlignBench v1.1, MTBench, Arena-Hard, and AlpacaEval. The full score of both AlimentBench and MTBench is 10 points. When calculating the average scores for alignment tasks, we converted the scores to a 100-point scale through linear mapping. 
    \item \textbf{Instruction following}: For instruction-following performance, we report accuracy of IFEval and score of SysBench, a benchmark that systematically analyzes system message following ability.
\end{itemize}

\subsubsection{Benchmark Results}
The Hunyuan-TurboS model demonstrates significant performance improvements through innovative methodologies including hybrid long-short chain-of-thought integration and multi-phase reinforcement learning. Our comprehensive evaluation  reveals the following key findings:

\textbf{Mathematical Reasoning}
 Hunyuan-TurboS achieves state-of-the-art performance among no-reasoning models, ranking second only to DeepSeek-v3-0324. The performance gap manifests primarily in AIME2024 and AIME2025 benchmarks (1-3 sample difference), while showing substantial advantages over other no-reasoning models across multiple datasets including GSM8k (hunyuan 94.4\% vs  gpt4.5 91.9\%), MATH (hunyuan 90\% vs gpt4.5 86.2\%), and OlympiadBench (hunyuan 76.1\% vs gpt4.5 67.6\%). 
 
\textbf{Logical Reasoning}
On complex benchmark suites (BBH, DROP, Zebra-Logic), Hunyuan-TurboS and DeepSeek-V3-0324 establish new performance plateaus, demonstrating substantial advantages over conventional no-reasoning models. 

\textbf{Coding}
Code generation capabilities are competitive with Qwen2.5-Max, showing equivalent performance to industry-leading models. The model shows significant advantages over  Doubao1.5-Pro-32k (+6.0) and  DeepSeek V3(+2.2) in programming tasks.

\textbf{Knowledge \& Chinese Tasks}
In terms of knowledge retention and factual accuracy, Hunyuan-TurboS achieves state-of-the-art (SOTA) performance comparable to leading foundation models. Benchmark evaluations reveal its robust capabilities in knowledge-intensive tasks, with particularly strong performance on Chinese-oriented knowledge assessments including C-Eval,  CMMLU, and C-SimpleQA. 

\begin{table*}\small  
\renewcommand\arraystretch{1.3}
\renewcommand\tabcolsep{1.6pt}
\centering
\begin{tabular}{l | c }
\hline

{Models} &  Average Output Tokens
 \\

\hline

DeepseekR1  
 &2283.5
 \\
Qwen3-235B-A22B   
 &2979.2
 \\
 \hline
Hunyuan-TurboS   
 &\textbf{1207.8}
 \\
\hline

\end{tabular}
\caption{\label{font-table}  The Results of token output from various API responses to STEM and general tasks. Hunyuan-TurboS achieves more cost-efficient output generation.
}
\label{result:table_5}
    \vspace{-1em}  
\end{table*}

\paragraph{Alignment Tasks}
Our evaluation demonstrates that Hunyuan-TurboS achieves significant performance improvements in alignment tasks, surpassing GPT-4.5 across multiple benchmarks. Specifically, Hunyuan-TurboS attains an average score 11.8 points higher than GPT-4.5 on AlpacaEval, while achieving state-of-the-art results with first-place rankings in both AlignmentBench and MTBench evaluations. 

\textbf{Instruction Following}
The model achieves state-of-the-art alignment performance, outperforming DeepSeek-V3-0324 by 5.0 points on IF-Eval. Its performance profile parallels Claude3.7 and GPT-4.5, with particularly strong results in multi-turn dialogue and constraint-based task execution.

\subsubsection{Inference efficiency of Adaptive CoT.} 

Efficient Inference is paramount for the practical application of LLMs, particularly in interactive scenarios that demand rapid responses and cost-effective operation. To assess the serving efficiency of Hunyuan-TurboS in comparison to other leading reasoning models, we conducted inference cost evaluations across a diverse set of tasks. The 6k evaluation dataset comprised 30\% STEM data and 70\% general domain data. To ensure statistical robustness and mitigate measurement bias, each inference request was executed five times, and the average values were subsequently reported for analysis.

As detailed in Table \ref{result:table_5} and Table \ref{tab:hunyuan_TurboS_LMSYS}, the results demonstrate that Hunyuan-TurboS achieves the most cost-efficient output generation among all evaluated models. Notably, our model delivers performance comparable to that of Deepseek-R1 in the LMSYS Chatbot Arena while utilizing only 52.8\% of the tokens, highlighting significant improvements in token efficiency which directly translate to reduced generation costs. Furthermore, Hunyuan-TurboS requires merely 40.5\% of Qwen3-235B-A22B's generation cost. These findings demonstrate the effectiveness of our proposed Adaptive Long-short Chain-of-Thought Fusion approach, also underscore Hunyuan-TurboS's strong capabilities in providing high-performance LLM inference with superior cost-effectiveness.

\begin{figure}[hbp]
    \centering
    \includegraphics[width=0.9\textwidth, trim=20 464 30 274,clip]{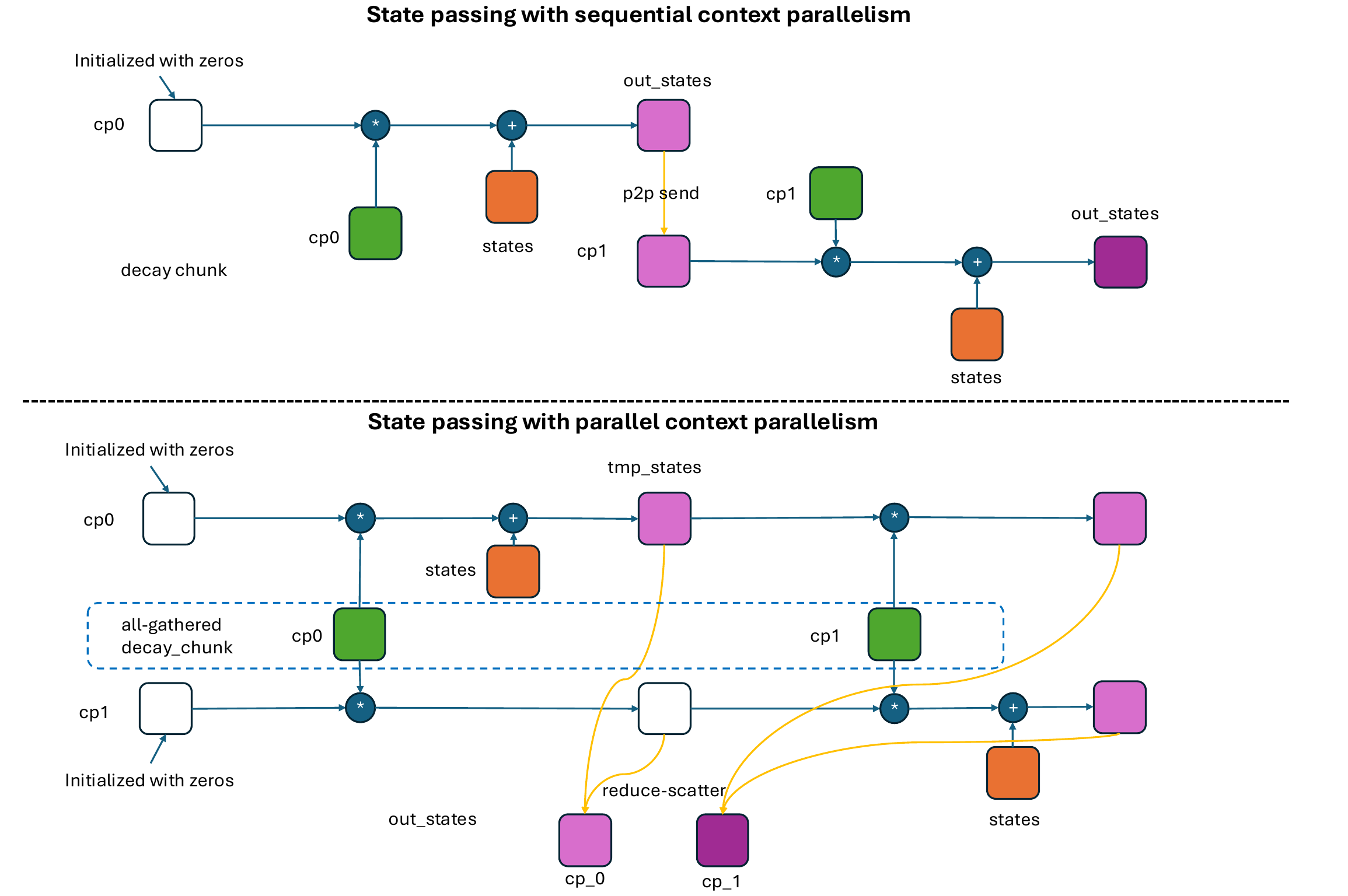}
    
    \caption{A diagram illustrating the Context Parallelism of Hunyuan-TurboS infrastructures.}
    \label{fig:Infra_fig_1}
\end{figure}

%% file: content/arch.tex
\section{Infrastructures}

\textbf{Reinforcement Training Framework.} The reinforcement training of Hunyuan-TurboS is based on Angel-RL, an efficient and lightweight reinforcement learning framework developed by Tencent that integrates both training and inference capabilities. Angel-RL is built upon Tencent's proprietary large-model training framework, AngelPTM~\citep{nie2023angel}, and the large-model inference framework, AngelHCF.

To achieve highly efficient reinforcement training for Hunyuan-TurboS, we have implemented meticulous engineering optimizations. Specifically, on the training side, we comprehensively integrate all model parallelism techniques and optimization strategies, including Tensor Parallelism (TP), Pipeline Parallelism (PP), Expert Parallelism (EP), Context Parallelism (CP), and sequence concatenation optimization to improve efficiency. As for context parallelism in particular, we implemented two state-passing approaches for context parallelism, namely the sequential and parallel paradigms, as illustrated in Figure~\ref{fig:Infra_fig_1}. The sequential implementation operates by having the preceding CP rank compute the final state, which is subsequently propagated to the next CP rank. The subsequent CP rank then employs this received state as the initial state to perform state passing computations. In contrast, the parallel approach first executes an all-gather operation on $decay_{chunk}$ within the CP group, enabling all CP ranks to concurrently conduct state passing computations. This parallel computation strategically omits calculations pertaining to $decay_{chunk} \times states$ that reside on non-local CP ranks. Ultimately, the output states generated through parallel computation undergo a reduce-scatter operation to yield the final consolidated results. On the sampling side, we support INT8 quantization~\citep{dettmers2022gpt3}.

Secondly, leveraging Tencent's custom-built Starlink Network, we effectively implement communication-computation overlap strategies, enabling a more seamless integration of communication and computation processes, thereby enhancing the overall system efficiency.

Furthermore, considering that reinforcement training often involves multiple large models for both training and inference, GPU memory typically becomes a bottleneck. To address this challenge, we have designed a multi-model reinforcement training workflow that combines hybrid and dedicated resource allocation. Additionally, using AngelPTM’s ZeroCache technology, we reduce the GPU memory pressure during large model reinforcement training by storing deduplicated model states and offloading them to CPU memory.

These innovations ensure that models with over 500 billion parameters can be efficiently trained within the Angel-RL framework. This not only expands the range of model sizes that can be handled but also advances the capabilities of large-scale model training in the field of artificial intelligence.

\textbf{Inference and Deployment.} 
Compared with prior Hunyuan-Turbo model,  Hunyuan-TurboS were designed for high efficiency and low latency at all context lengths. The inference of the Hunyuan-TurboS model is powered by the AngelHCF Inference Acceleration Framework. For the Mamba Hybrid architecture of the TurboS model, we have implemented optimizations across folloing three key dimensions, ultimately achieving a 1.8x speedup compared to Hunyuan-Turbo, which is a pure Transformers MoE model:

\begin{enumerate}[label=(\arabic*)]

    \item \textbf{Mamba Kernel Optimization}: 
        \begin{itemize}
        \item \textbf{Prefill Phase}: By exploiting the structural features of Mamba2, we aimed to augment computational parallelism. The workload was partitioned into several parallelizable General Matrix Multiply (GEMM) operations, and a chunkscan block-wise parallel computation strategy was implemented.
        \item \textbf{Decode Phase}: In order to mitigate memory bandwidth limitations, the SelectivescanUpdate kernel was devised. This kernel facilitates more accurate read and update operations on Mamba state vectors at specified locations.
    \end{itemize}
    \item \textbf{MoE Optimization}: Beyond conventional Tensor Parallel partitioning, we prioritized Expert Parallel to mitigate memory bottlenecks during decoding. We employed intelligent redundant expert allocation to balance computational load across GPUs. In collaboration with the networking team, we optimized communication for both dispatch and combine phases while implementing computation-communication overlap, achieving significant throughput improvements.
    
    \item \textbf{Hybrid Architecture Precision Optimization}: While Mamba’s linear structure offers computational efficiency advantages over attention mechanisms in long-context scenarios, its inherent lack of global attention capture can lead to repetitive generation issues when using standard fp16/bf16 for Mamba state. To address this, we innovatively adopted fp32 precision for Mamba state at the kernel level, elevating the long-text generation quality of the hybrid architecture to match that of Full Attention models. This optimization reduced token consumption by 35\%-45\% in mathematically intensive and programming competition-level reasoning tasks (compared to the original fp16/bf16 approach).
 
\end{enumerate}

%% file: content/conclusion.tex
\section{Conclusion}
\label{sec:conclusion}

In this report, we introduced Hunyuan-TurboS, a novel large hybrid Transformer-Mamba Mixture-of-Experts (MoE) model. It uniquely synergizes Mamba's long-sequence processing efficiency with Transformer's superior contextual understanding, incorporating an innovative AMF/MF block pattern and an adaptive long-short Chain-of-Thought (CoT) mechanism. Pre-trained on 16T high-quality tokens and supporting a 256K context length, this 56B activated parameter (560B total) model stands as the first industry-deployed large-scale Mamba architecture. Our comprehensive post-training regimen, featuring Supervised Fine-Tuning, Adaptive Long-short CoT Fusion, Multi-round Deliberation Learning, and a two-stage Reinforcement Learning process, significantly enhanced its capabilities. Hunyuan-TurboS demonstrates strong performance, achieving a 1356 LMSYS Chatbot Arena score and averaging 77.9\% across 23 automated benchmarks. Critically, Hunyuan-TurboS strikes an effective balance between high performance and computational efficiency, delivering substantial capabilities at lower inference costs than many reasoning models. This work establishes a new paradigm for efficient, large-scale pre-trained models, advancing the development of accessible and powerful AI systems.

%% file: content/authors.tex
\newpage
\section{Authors}
Within each role, authors are listed alphabetically.

\begin{multicols}{2} %

\noindent
\textbf{Core Contributors} \\
Ao Liu \\ 
Botong Zhou \\ 
Can Xu \\ 
Chayse Zhou \\ 
ChenChen Zhang \\ 
Chengcheng Xu \\ 
Chenhao Wang \\ 
Decheng Wu \\ 
Dengpeng Wu \\ 
Dian Jiao \\ 
Dong Du \\ 
Dong Wang \\ 
Feng Zhang \\ 
Fengzong Lian \\ 
Guanghui Xu \\ 
Guanwei Zhang \\ 
Hai Wang \\ 
Haipeng Luo \\ 
Han Hu \\ 
Huilin Xu \\ 
Jiajia Wu \\ 
Jianchen Zhu \\ 
Jianfeng Yan \\ 
Jiaqi Zhu \\ 
Jihong Zhang \\ 
Jinbao Xue \\ 
Jun Xia \\ 
Junqiang Zheng \\ 
Kai Liu \\ 
Kai Zhang \\ 
Kai Zheng \\ 
Kejiao Li \\ 
Keyao Wang \\ 
Lan Jiang \\ 
Lixin Liu \\ 
Lulu Wu \\ 
Mengyuan Huang \\ 
Peijie Yu \\ 
Peiqi Wang \\ 
Qian Wang \\ 
Qianbiao Xiang \\ 
Qibin Liu \\ 
Qingfeng Sun \\ 
Richard Guo \\ 
Ruobing Xie \\ 
Saiyong Yang \\ 
Shaohua Chen \\ 
Shihui Hu \\ 
Shuai Li \\ 
Shuaipeng Li \\ 
Shuang Chen \\ 
Suncong Zheng \\ 
Tao Yang \\ 
Tian Zhang \\ 
Tinghao Yu \\ 
Weidong Han \\ 
Weijie Liu \\ 
Weijin Zhou \\ 
Weikang Wang \\ 
Wesleye Chen \\ 
Xiao Feng \\ 
Xiaoqin Ren \\ 
Xingwu Sun \\ 
Xiong Kuang \\ 
Xuemeng Huang \\ 
Xun Cao \\ 
Yanfeng Chen \\ 
Yang Du \\ 
Zhen Yang \\ 
Yangyu Tao \\ 
Yaping Deng \\ 
Yi Shen \\ 
Yigeng Hong \\ 
Yiqi Chen \\ 
Yiqing Huang \\ 
Yuchi Deng \\ 
Yue Mao \\ 
Yulong Wang \\ 
Yuyuan Zeng \\ 
Zenan Xu \\ 
Zhanhui Kang \\ 
Zhe Zhao \\ 
ZhenXiang Yan \\ 
Zheng Fang \\ 
Zhichao Hu \\ 
Zhongzhi Chen \\ 
Zhuoyu Li \\ 
Zongwei Li \\

\noindent
\textbf{Contributors} \\
Alex Yan \\
Ande Liang \\ 
Baitong Liu \\ 
Beiping Pan \\ 
Bin Xing \\ 
Binghong Wu \\ 
Bingxin Qu \\ 
Bolin Ni \\ 
Boyu Wu \\ 
Chen Li \\ 
Cheng Jiang \\ 
Cheng Zhang \\ 
Chengjun Liu \\ 
Chengxu Yang \\ 
Chengzhong Xu \\
Chiyu Wang \\ 
Chong Zha \\ 
Daisy Yi \\ 
Di Wang \\ 
Fanyang Lu \\ 
Fei Chen \\ 
Feifei Liu \\ 
Feng Zheng \\ 
Guanghua Yu \\ 
Guiyang Li \\ 
Guohua Wang \\ 
Haisheng Lin \\ 
Han Liu \\ 
Han Wang \\ 
Hao Fei \\ 
Hao Lu \\ 
Haoqing Jiang \\ 
Haoran Sun \\ 
Haotian Zhu \\ 
Huangjin Dai \\ 
Huankui Chen \\ 
Huawen Feng \\ 
Huihui Cai \\ 
Huxin Peng \\ 
Jackson Lv \\ 
Jiacheng Shi \\ 
Jiahao Bu \\ 
Jianbo Li \\ 
Jianglu Hu \\ 
Jiangtao Guan \\ 
Jianing Xu \\ 
Jianwei Cai \\ 
Jiarong Zhang \\ 
Jiawei Song \\ 
Jie Jiang \\ 
Jie Liu \\ 
Jieneng Yang \\ 
Jihong Zhang \\ 
Jin lv \\ 
Jing Zhao \\ 
Jinjian Li \\ 
Jinxing Liu \\ 
Jun Zhao \\ 
Juntao Guo \\ 
Kai Wang \\ 
Kan Wu \\ 
Lei Fu \\ 
Lei He \\ 
Lei Wang \\ 
Li Liu \\ 
Liang Dong \\ 
Liya Zhan \\ 
Long Cheng \\ 
Long Xu \\ 
Mao Zheng \\ 
Meng Liu \\ 
Mengkang Hu \\ 
Nanli Chen \\ 
Peirui Chen \\ 
Peng He \\ 
Pengju Pan \\ 
Pengzhi Wei \\ 
Qi Yang \\ 
Qi Yi \\ 
Roberts Wang \\ 
Rongpeng Chen \\ 
Rui Sun \\ 
Rui Yang \\ 
Ruibin Chen \\ 
Ruixu Zhou \\ 
Shaofeng Zhang \\ 
Sheng Zhang \\ 
Shihao Xu \\ 
Shuaishuai Chang \\ 
Shulin Liu \\ 
SiQi Wang \\ 
Songjia  Feng \\ 
Songling Yuan \\ 
Tao Zhang \\ 
Tianjiao Lang \\ 
Tongkai Li \\ 
Wei Deng \\ 
Wei Li \\ 
Weichao Wang \\ 
Weigang Zhang \\ 
Weixuan Sun \\ 
Wen Ouyang \\ 
Wenxiang Jiao \\ 
Wenzhi Sun \\ 
Wenzhuo Jia \\ 
Xiang Zhang \\ 
Xiangyu He \\ 
Xianshun Ren \\ 
XiaoYing Zhu \\ 
Xiaolong Guo \\ 
Xiaoxue Li \\ 
Xiaoyu Ma \\ 
Xican Lu \\ 
Xinhua Feng \\ 
Xinting Huang \\ 
Xinyu Guan \\ 
Xirui Li \\ 
Xu Zhang \\ 
Xudong Gao \\ 
Xun Luo \\ 
Xuxiang Qi \\ 
Yangkun Chen \\ 
Yangyu Tao \\ 
Yanling Xiao \\ 
Yantao Mai \\ 
Yanze Chen \\ 
Yao Ding \\ 
Yeting Yang \\ 
YiFan Song \\ 
Yifan Yang \\ 
Yijiao Zhu \\ 
Yinhe Wu \\ 
Yixian Liu \\ 
Yong Yang \\ 
Yuanjun Cai \\ 
Yuanlin Tu \\ 
Yue Zhang \\ 
Yufei Huang \\ 
Yuhang Zhou \\ 
Yuhao Jiang \\ 
Yuhong Liu \\ 
Yuhui Hu \\ 
Yujin Lin \\ 
Yun Yang \\ 
Yunhao Wang \\ 
Yusong Zhang \\ 
Zekun Wu \\ 
Zelong Zhang \\ 
Zhan Yu \\ 
Zhaoliang Yang \\ 
Zhe Zhao \\ 
Zheng Li \\ 
Zhenyu Huang \\ 
Zhiguang Liu \\ 
Zhijiang Xu \\ 
Zhiqing Kui \\ 
Zhiyin Zeng \\ 
Zhiyuan Xiong \\ 
Zhuo Han \\ 
Zifan Wu \\ 
Zigang Geng \\ 
Zilong Zhao \\ 
Ziyan Tang \\ 
Ziyuan Zhu \\ 
Zonglei Zhu \\ 
Zhijiang Xu \\

\end{multicols} %

%% file: content/appendix.tex
\section{Appendix}
\label{sec:appendix}
\begin{figure}[!htb]
\centering
     \includegraphics[width=0.8\textwidth, scale=1, trim=0 175 0 0 ,clip]{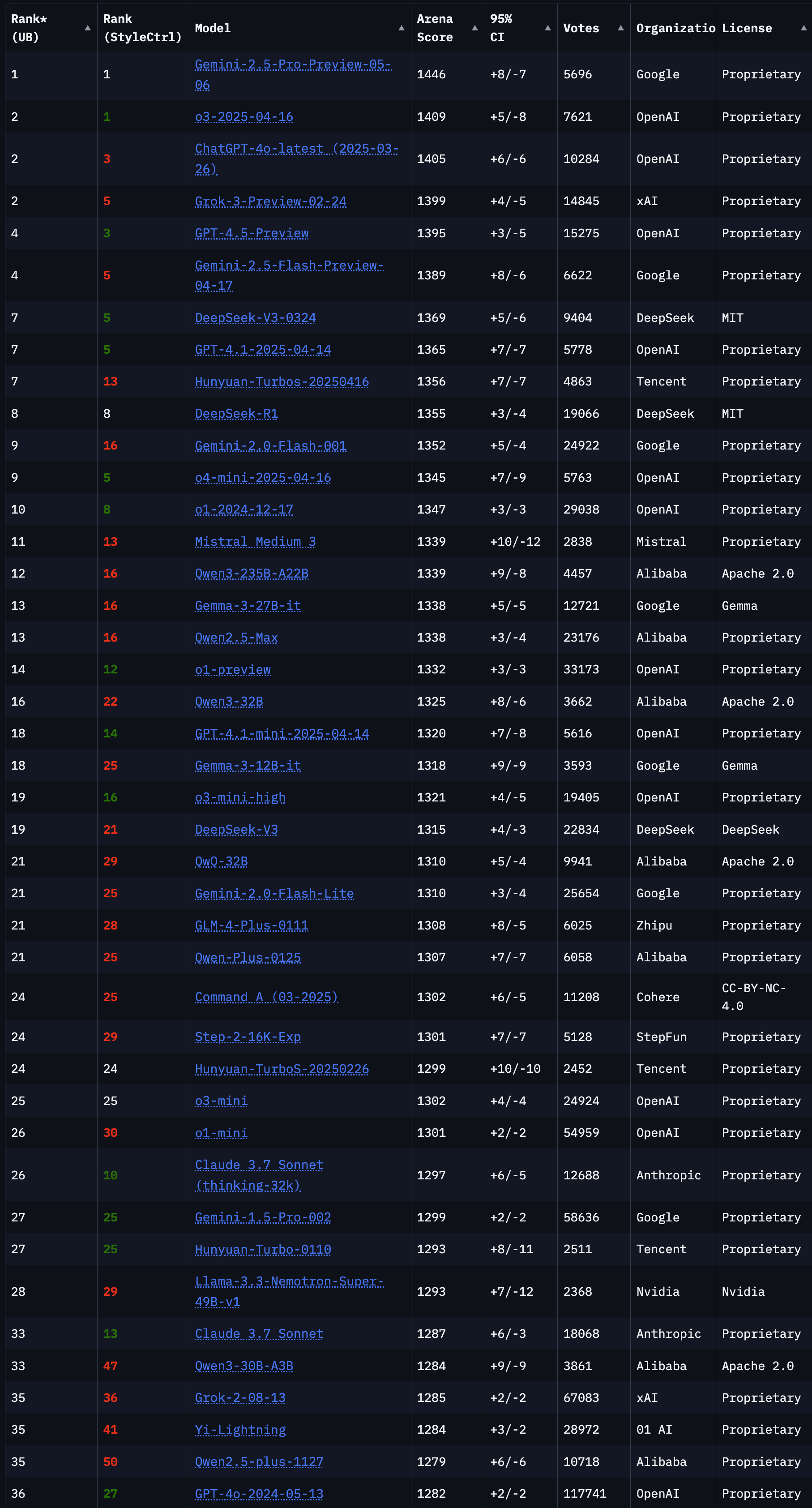}
     \caption{Full Leaderboard of LMSYS Chatbot Arena by May 18, 2025.}
     \label{fig:apdx_figure1}
\end{figure}

Figure \ref{fig:apdx_figure1} present the full result of  Leaderboard of LMSYS Chatbot Arena by May 18, 2025.
